\documentclass{article}
\usepackage{arxiv}

\usepackage[utf8]{inputenc} 
\usepackage[T1]{fontenc}    
\usepackage{hyperref}       
\usepackage{url}            
\usepackage{booktabs}       
\usepackage{amsfonts}       
\usepackage{nicefrac}       
\usepackage{microtype}      
\usepackage{lipsum}		
\usepackage{graphicx}
\usepackage{natbib}
\usepackage{doi}
\usepackage{amsmath}
\usepackage{amssymb}
\usepackage{booktabs}
\usepackage{multirow}
\usepackage{bbding}
\usepackage{pifont}
\usepackage{graphicx}    
\usepackage{subcaption}  

\usepackage{color}

\usepackage[table]{xcolor}

\definecolor{C1}{HTML}{93BFCF}
\definecolor{C2}{HTML}{A0C3D2}
\definecolor{C3}{HTML}{BDCDD6}
\definecolor{C4}{HTML}{EEE9DA}
\definecolor{C5}{HTML}{FFF1DC}
\definecolor{C6}{HTML}{E8D5C4}
\definecolor{C7}{HTML}{EEEEEE}
\definecolor{C8}{HTML}{BCEE68}

\title{Gaussian2Scene: 3D Scene Representation Learning via Self-supervised Learning with 3D Gaussian Splatting}

\author{Keyi Liu$^1$ , Weidong Yang$^{1,}$\Envelope, Ben Fei$^{2,}$\Envelope, Ying He$^{3}$\\
$^1$ Fudan University, $^2$ The Chinese University of Hong Kong, $^3$ Nanyang Technological University\\
\texttt{23210240242.m.fudan.edu.cn, wdyang@fudan.edu.cn, benfei@cuhk.edu.hk}\\
\Envelope Corresponding Authors
}

\date{}

\renewcommand{\shorttitle}{Gaussian2Scene}

\hypersetup{
pdftitle={Gaussian2Scene: 3D Scene Representation Learning via Self-supervised Learning with 3D Gaussian Splatting},
pdfsubject={},
pdfauthor={},
pdfkeywords={},
}

\begin{document}

\maketitle

\begin{abstract}

Self-supervised learning (SSL) for point cloud pre-training has become a cornerstone for many 3D vision tasks, enabling effective learning from large-scale unannotated data. 
At the scene level, existing SSL methods often incorporate volume rendering into the pre-training framework, using RGB-D images as reconstruction signals to facilitate cross-modal learning. This strategy promotes alignment between 2D and 3D modalities and enables the model to benefit from rich visual cues in the RGB-D inputs. However, these approaches are limited by their reliance on implicit scene representations and high memory demands. Furthermore, since their reconstruction objectives are applied only in 2D space, they often fail to capture underlying 3D geometric structures.
To address these challenges, we propose Gaussian2Scene, a novel scene-level SSL framework that leverages the efficiency and explicit nature of 3D Gaussian Splatting (3DGS) for pre-training. 
The use of 3DGS not only alleviates the computational burden associated with volume rendering but also supports direct 3D scene reconstruction, thereby enhancing the geometric understanding of the backbone network.
our approach follows a progressive two-stage training strategy. In the first stage, a dual-branch masked autoencoder learns both 2D and 3D scene representations. In the second stage, we initialize training with reconstructed point clouds and further supervise learning using the geometric locations of Gaussian primitives and rendered RGB images. This process reinforces both geometric and cross-modal learning.
We demonstrate the effectiveness of Gaussian2Scene across several downstream 3D object detection tasks, showing consistent improvements over existing pre-training methods.
\end{abstract}  

\section{Introduction}
\label{sec:instruction}


Deep neural networks have recently extended from 2D to 3D domains, demonstrating a vital role in real-world applications such as virtual reality, autonomous driving, and robotics~\citep{fei2022comprehensive}. 
Many of these applications require a well-trained feature encoder with advanced 3D understanding capability, especially in the recently evolving field of 3D Visual-Language-Action (3D-VLA)~\citep{ding2024quar} and Visual Language Navigation (VLN), where agents need to understand and recognize diverse visual scenes and align the extracted 3D features with human language instructions~\citep{long2024discuss}. 
As a common 3D data representation, point clouds provide rich location information and attribute information, making them widely used for training such 3D encoders.
However, despite the ease of acquiring raw point cloud data from sensors like LiDAR, their irregularity, sparsity, and possible loss and occlusion pose significant challenges for annotation—especially in complex scenes containing hundreds of thousands of points and numerous object categories~\citep{fei2024curriculumformer}.
In this case, self-supervised learning (SSL) reduces the need for annotations by leveraging unlabeled point cloud data for pre-training~\citep{fei2023self}.
Thus, the downstream tasks based on pre-trained models require only a small amount of labeled data, which can lead to excellent performance after fine-tuning~\citep{fei2024parameter}.\par
Current paradigms of self-supervised learning methods for point clouds can be generally classified into two categories: generative-based~\citep{wang2021unsupervised,yu2022point,pang2022masked,zhang2022point} and contrastive-based~\citep{xie2020pointcontrast,afham2022crosspoint,huang2021spatio}.
Generative-based methods typically design reconstruction tasks, enabling the network to learn geometric features from incomplete data and reconstruct point clouds from masked inputs~\citep{wang2021unsupervised,yu2022point,pang2022masked,zhang2022point}.
However, at the scene level, due to the irregularity and occlusion of point clouds limit the effectiveness of reconstruction alone, often resulting in imcomplete 3D feature learning.
Contrastive-based methods, on the other hand, aim to learn invariant representations under various geometric transformations~\citep{xie2020pointcontrast,afham2022crosspoint,huang2021spatio}. 
However, they face challenges in effectively aligning features due to the lack of positive and negative samples, as well as relying on simplistic data augmentation strategies.\par
Unlike these methods, Huang et al.~\citep{huang2023ponder} proposed volume rendering-based methods for point cloud self-supervised pre-training. 
The differentiable rendering decoder takes an implicitly encoded 3D feature volume as input and outputs rendered color images and depth maps, which are supervised by ground truth images. 
However, the implicit volume representation requires high computational and memory demands for optimization, limiting its efficiency~\citep{song2024city}.
Additionally, since their reconstruction objectives are applied only in 2D space, these methods often fail to capture underlying 3D geometric structures.
Recently, 3D Gaussian Splatting (3DGS) has introduced an explicit representation of 3D data, achieving real-time rendering speed and high-quality novel view synthesis~\citep{kerbl20233d,fei20243d}.
Inspired by this novel scene representation method, we leverage 3D Gaussian splatting in point cloud self-supervised learning pretraining. 
This approach not only provides RGB images through real-time differentiable rasterization but also incorporates 3D Gaussian anchors as geometric information.

Inspired by previous pre-training methods that fuse point clouds with images through rendering~\citep{liu2025gs}, we propose Gaussian2Scene, a framework that fully utilizes the powerful rendering and representation capabilities of 3DGS  to achieve a deep understanding of the scene through a cross-modal two-stage pre-training paradigm.
Specifically, in the first stage, the model processes the data of the 2D images and 3D point clouds based on two MAE architectures. 
Each branch learns the modality-specific features through an independent reconstruction task. 
The spatial coordinates of the point cloud can be aligned with the pixel plane of the image via camera projection. 
Subsequently, the model further fuses the features of the two modalities through a shared Transformer structure with cross-modal attention, enabling the network to learn both geometric and color information in an integrated manner. 
In the second stage, we introduced 3DGS as a differentiable renderer. 
The point cloud reconstructed from the point branch is used as the initialization position of the Gaussian primitive. 
Through the differentiable rendering of 3DGS, the network is able to generate high-quality rendered images and extract richer position details from optimized Gaussians. 
This process allows the model to progressively enhance its understanding of scene structure and appearance, further integrating multimodal information to capture more comprehensive and detailed scene representations.
During pre-training, Gaussian2Scene achieves geometric alignment between Gaussian-optimized point clouds and original inputs while rendering high quality images at a average of 26.4 PSNR.
When transferred to 3D object detection via 3DETR~\citep{misra2021end}, our approach achieves $AP_{50}$ gains of $33.5\%$ on SUN RGB-D~\citep{song2015sun} and $43.3\%$ on ScanNetV2~\citep{dai2017scannet} and maintains  $AP_{25}$ of  $59.2\%$ and $62.9\%$ on both datasets.
These results collectively demonstrate that explicit Gaussian representations provide geometrically superior supervision signals for transferable 3D understanding and the effectiveness of the multi-modal supervision strategy in learning transferable representations.

Our contributions can be summarized as follows:

\begin{itemize}
    \item We propose \textbf{Gaussian2Scene}, a novel scene-level self-supervised pre-training framework that leverages the explicit and efficient 3D Gaussian Splatting representation to address the limitations of implicit volume rendering-based methods. Our approach  enables direct 3D scene reconstruction, enhancing geometric understanding of the 3D backbone.
    \item We introduce a progressive two-stage training strategy that combines modality-specific and cross-modal learning. The first stage employs a dual-branch masked autoencoder to jointly learn 2D image and 3D point cloud features. The second stage initializes with reconstructed point clouds and further refines learning via supervision on 3DGS primitive geometry and rendered RGB images, facilitating end-to-end multimodal optimization.
    \item We devise 3DGS-aware pre-training pipeline that integrates differentiable rendering and geometric consistency losses. By optimizing anisotropic 3D Gaussians with learnable parameters, our framework explicitly models scene geometry while maintaining alignment between 2D and 3D modalities through joint supervision.
\end{itemize}

\section{Related Work}
\label{sec:related}

\subsection{3D Gaussian Splatting}
Recently, 3DGS has gained significant advancements, attributed primarily to its remarkable rendering speed and ability to synthesize realistic scenes from novel perspectives.
Compared with Neural Radiance Fields (NeRF)~\citep{mildenhall2021nerf}, 3DGS introduces a more compact and efficient representation by utilizing Gaussian distributions to model the uncertainty and density of 3D points.
Therefore, 3DGS has been widely used for surface reconstruction~\citep{guedon2024sugar}, dynamic modeling~\citep{yang2024deformable}, large-scene modeling~\citep{lin2024vastgaussian}, scene manipulation~\citep{chen2024gaussianeditor}, 3D generation~\citep{liang2024luciddreamer}, 3D perception~\citep{zhou2024drivinggaussian} and human modeling~\citep{jiang2024hifi4g}.
However, utilizing 3DGS for point cloud self-supervised learning is still an under-explored area.
\subsection{Self-supervised Learning in Point Clouds}
Several methodologies~\citep{fei2024curriculumformer,fei2023self} have been developed and examined for self-supervised learning on point clouds. Generally, existing methods can be categorized into contrastive methods and generative methods.

\textbf{Generative Methods} employ an encoder-decoder architecture to learn representations from point cloud via self-reconstruction~\citep{wang2021unsupervised,yu2022point,pang2022masked,zhang2022point}.
Inspired by the success of the masked auto-encoder (MAE) in 2D computer vision, Pang et al.~\citep{pang2022masked} proposed Point-MAE for 3D point clouds. 
Point-M2AE~\citep{zhang2022point} advances upon Point-MAE by addressing the limitations related to encoding single-resolution point clouds and neglection of local-global relations in 3D shapes.
Through skip connections between encoder and decoder stages, Point-M2AE enhances fine-grained information during up-sampling, promoting local-to-global reconstruction and capturing the relationship between local structure and global shape.

\textbf{Contrastive Methods} learn discriminative features by training network to distinguish between positive and negative samples~\citep{xie2020pointcontrast,afham2022crosspoint,huang2021spatio}.
CrossPoint~\citep{afham2022crosspoint} is a cross-modal contrastive learning method,     which introduces a contrastive loss between the rendered 2D image feature and the point cloud feature.

\subsection{Scene-level 3D Self-supervised Learning}
The object-level point cloud pre-training methods described above typically involve 3D shapes data. In contrast, some recent studies focus on scene-level point cloud pre-training. 
In this context, networks trained on single-modality point cloud exhibit limited capacity to learn comprehensive scene representations. To address this, rendering is utilized to be a potent technique for enhancing 3D encoders.
Ponder~\citep{huang2023ponder}~\citep{zhu2023ponderv2} designs a sparse point cloud encoder within a volumn-based rendering decoder, where depth and color are parameterized along camera rays and predicted by MLPs. The network is trained by minimizing the difference between rendered and ground truth images.
CluRender~\citep{mei2024unsupervised} leverages neural rendering after point soft-clustering encoder to learn cross-modal features in an implicit manner. 
Similarly, render-based self-supervised learning frameworks have made progress in outdoor scenes. UniPAD~\citep{yang2024unipad} extends this paradigm to outdoor autonomous driving by introducing a neural rendering decoder that reconstructs masked regions using depth-aware sampling and ray integration.
Different from NeRF-based work, we utilize 3DGS as the bridge of 2D images and 3D representaion, and capitalize both on the high quality rendering ability of differentiable Gaussian splatting in 2D and the explict parametric 3D information of Gaussian primitives. 
%


\section{Gaussian2Scene}
\label{sec:method}
\begin{figure*}[t]                   
    \centering  
    \includegraphics[width=\textwidth]{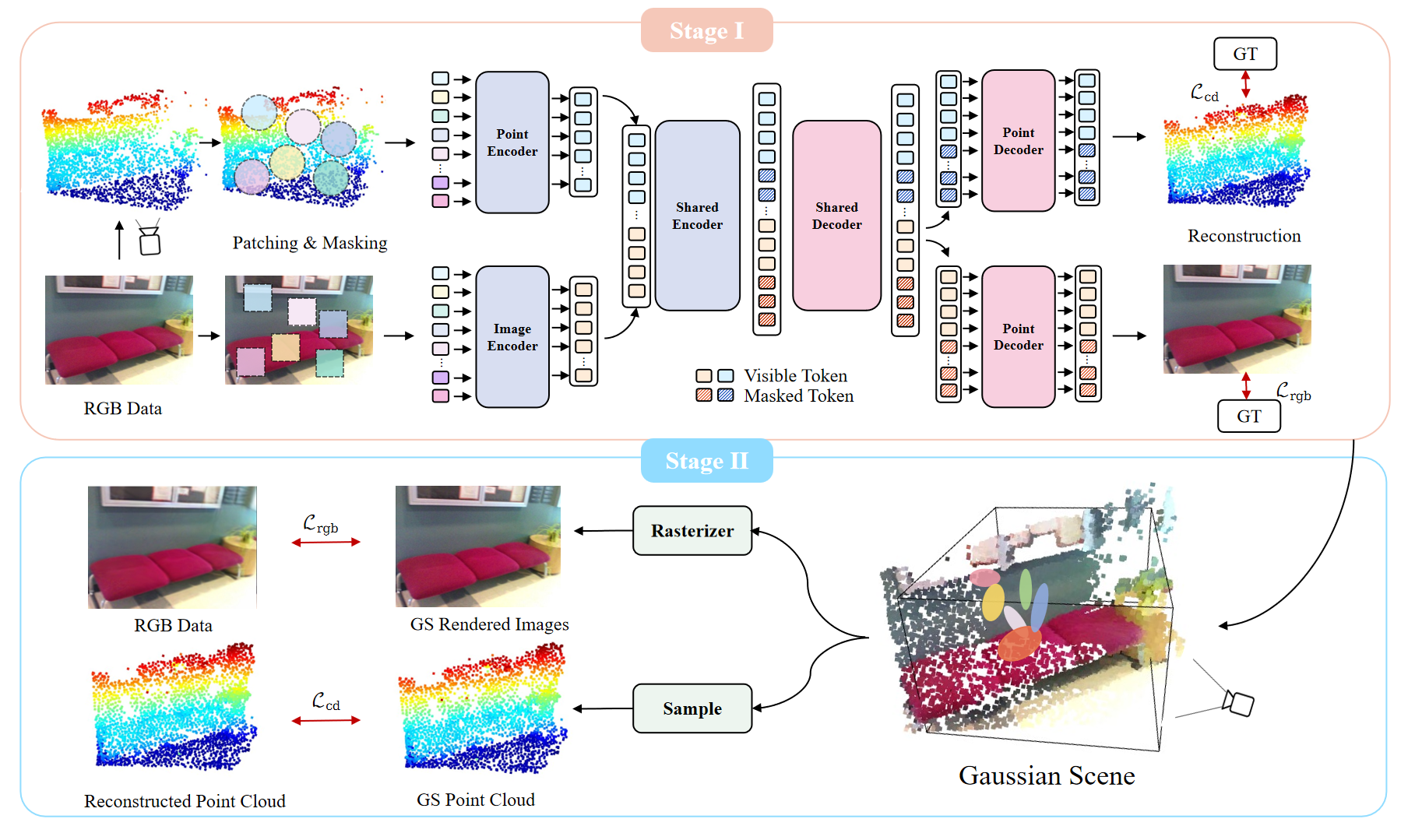} 
    \caption{The pipeline of Gaussian2Scene follows a progressive two-stage training strategy. In the first stage, the cross-modal MAE-based networks are used to learn and fuse 2D and 3D features of the scene. In the second stage, it leverages the rendering capabilities of 3DGS to apply 2D reconstruction loss on rendered points and to extract geometric information from Gaussian primitives.}  
    \label{fig:pipeline}  
\end{figure*}  
\begin{figure*}[t]                   
    \centering  
    \includegraphics[width=\textwidth]{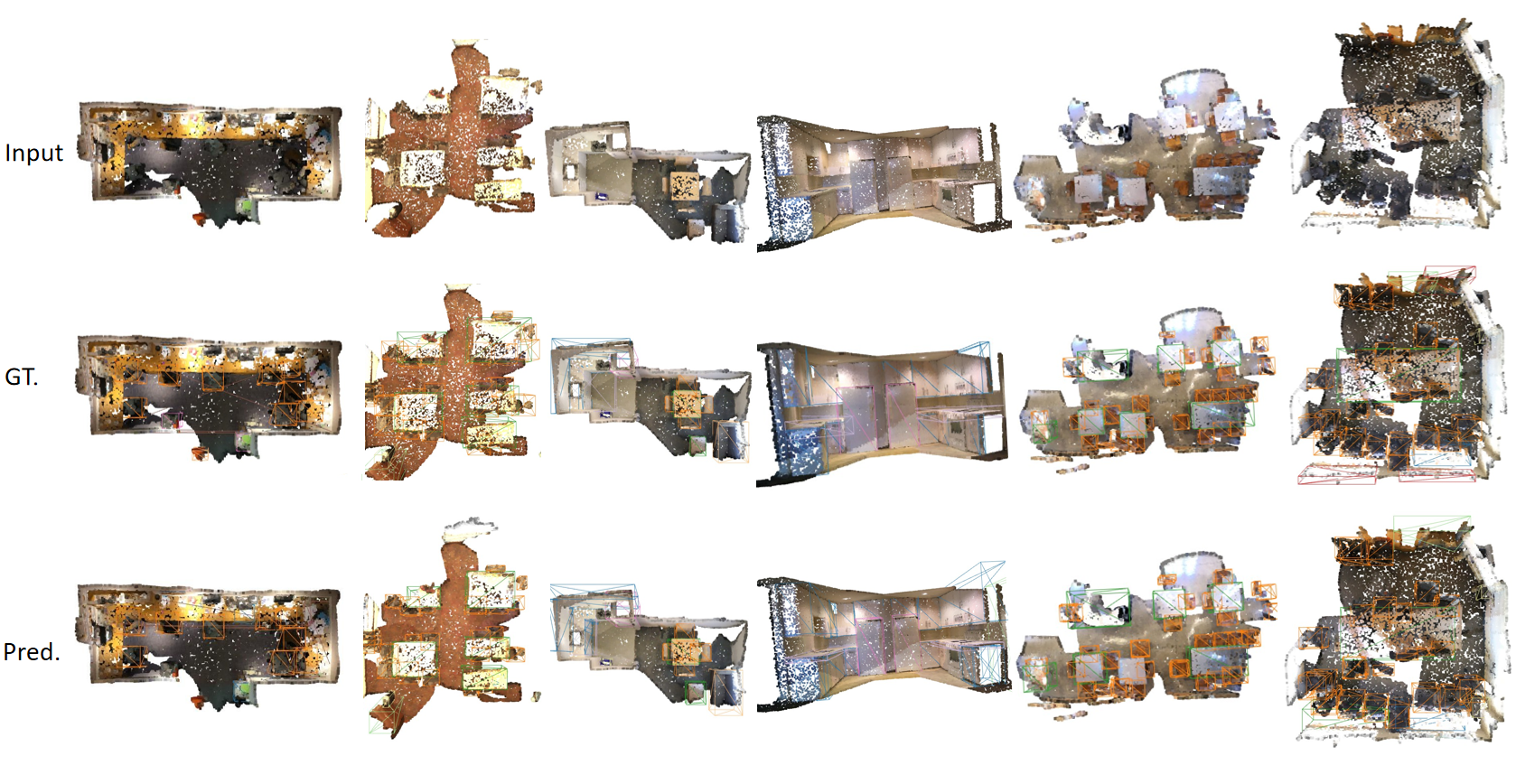} 
    \caption{The downstream object detection results on ScanNetV2~\cite{dai2017scannet}.}  
    \label{fig:detection vis}  
\end{figure*}  
In this section, we will introduce our progressive two-stage pre-training method.
Firstly, we pre-train a dual-branch encoder-decoder with mask and reconstruction tasks of point cloud and image, separately. Secondly, we introduce 3DGS into the framework, learning features from 2D and 3D modality.
The pipeline of Gaussian2Scene is shown in Figure~\ref {fig:pipeline}.

\subsection{Masked Autoencoding Pre-training}

In the first stage of pre-training, inspired by ~\citep{chen2023pimae}, we use a two-branch MAE learning framework that jointly learns the cross-modal features of both point cloud and corresponding color images.

\subsubsection{Cross-modal Modules}
The two encoders from the image and point cloud branches take visible tokens with their positional and modality embeddings as input to learn the representations of the features, separately. 
The point branch is based on ~\citep{zhang2022point}. 
The input point clouds are first processed into multiple local patches and embedded into cluster tokens using the Farthest Point Sampling (FPS) and the k-nearest neighbor (kNN) algorithms.
Specifically, the input point clouds $\mathcal{P} \in \mathbb{R}^{N \times 3}$ are partitioned into $M$ local patches using FPS-kNN clustering ($M=64$, $k=32$),  embedded into visible tokens $\mathbf{F}_p \in \mathbb{R}^{M \times C}$ using a lightweight PointNet and combined with learnable positional embeddings  $\mathbf{E}_{pos}$. 
A random masking ratio 60\% is used, where the corrupted patches are replaced by mask tokens $E_{mask}$. 
With the features learned through the encoder and shared modules, the point cloud decoder is used to decode the high-level latent representation into points. 
Chamfer Distance $L_{\text{point-rec}}$ is utilized for loss calculation after reconstruction.\par
For the image branch, the 3D coordinates of the point cloud are projected onto the 2D image plane through the internal and external parameters of the camera. Following ~\citep{chen2023pimae}, complementary masking is achieved after aligning the point cloud token with the image block.
Image Encoder, following ViT~\citep{dosovitskiy2020image}, take visible tokens with their respective positional and modality embeddings as input.
Image decoder finally takes the features seperated from results of shared module and reconstructs the image using MSE loss $L_{\text{image-rec}}$.

\subsubsection{Fusion of Modalities}
A shared encoder-decoder fuses the aligned latent features of two modalities to generate cross-modal representations. Specifically, image tokens and point cloud tokens are concatenated together, and cross-modal interactions are performed via the shared encoder-decoder. 
The attention mechanism and  complementary masks enable image tokens and point cloud tokens to attend to each other, facilitating comprehensive information fusion.
The output of the shared decoder is then split back into point cloud and image tokens, which serve as input to the subsequent modality-specific decoders.
The masked and embedded tokens are concatenated and fed into decoders, which reconstruct the corrupted patches.
For cross-modal reconstruction , the masked point cloud token is processed through a prediction head to estimate the corresponding image feature, with a loss using MSE loss between the predicted features and the real image features $L_{\text{cross-rec}}$.
The total loss is the sum of the seperated and joint loss terms, formulated as $L_{stage1}$:
\begin{equation}
L_{\text{stage1}} = L_{\text{point-rec}} +  L_{\text{image-rec}} +  L_{\text{cross-rec}} 
\end{equation}

\subsection{Pre-training with 3DGS Rendering}
The key contribution of the second stage of the pre-training approach leverages 3DGS to reconstruct radiance fields from multi-view images. 
The method represents scenes as a collection of anisotropic 3D Gaussians with learnable parameters, optimized through differentiable rendering. Leveraging the differentiable rendering capability of 3DGS, the rendering process can be integrated into the proposed pretraining framework, jointly optimizing both 3D and 2D losses to enhance overall performance.
The geometric information and the rendering results of the optimized scene are used as the additional supervision signals for pre-training.

\subsubsection{3DGS Representation}
The fundamental scene representation of 3DGS comprises $N$ anisotropic 3D Gaussians $\mathcal{G}_i = (\boldsymbol{\mu}_i, \boldsymbol{\Sigma}_i, \boldsymbol{c}_i, \alpha_i)$, where $\boldsymbol{\mu}_i  $ denotes the Gaussian center position, $\boldsymbol{\Sigma}_i $ represents the covariance matrix controlling spatial extent, $\boldsymbol{c}_i \in \mathbb{R}^3$ specifies the color, and $\alpha_i \in [0,1]$ determines opacity. To maintain positive semi-definiteness during optimization, the covariance matrix is decomposed into rotational $\boldsymbol{R}_i$ and scaling ${\boldsymbol{S}_i}$ components:
\begin{equation}
    \boldsymbol{\Sigma}_i = \boldsymbol{R}_i\boldsymbol{S}_i\boldsymbol{S}_i^\top\boldsymbol{R}_i^\top
\end{equation}
where $\boldsymbol{R}_i$ corresponds to a rotation matrix parameterized by quaternions, and $\boldsymbol{S}_i$ denotes a diagonal scaling matrix. This decomposition enables independent optimization of orientation and scale while maintaining mathematical validity. All the properties are learnable and optimized through back-propagation.

\subsubsection{Optimization and Rendering}
The initialization of GS begins with seeding Gaussians from reconstructed  point clouds from the decoder of point cloud branch. 3DGS utilize differentiable rendering to project 3D Gaussians to 2D image planes. The projection transform derives from the viewing transformation $\boldsymbol{W}$ and Jacobian $\boldsymbol{J}$ of the affine approximation:
\begin{equation}
    \boldsymbol{\Sigma}'_i = \boldsymbol{JW\Sigma}_i\boldsymbol{W}^\top\boldsymbol{J}^\top
\end{equation}
The properties of a 3D Gaussian can be optimized directly through back-propagation. After rendering image $I_{GS}$, supervised with the image $I_{\text{gt}}$ and the view of the training camera from the image branch, photometric optimization minimizes a composite loss function that combines $\mathcal{L}_1$, D-SSIM metrics~\citep{wang2004image}, and volume regularization:
\begin{equation}
\mathcal{L} = {\mathcal{L}_{\text{L1}}} + {\lambda_{\text{ssim}} \big(1 - \text{SSIM}(I_{GS}, I_{\text{gt}})\big)} + {\gamma \cdot \frac{1}{N} \sum_{i=1}^N \prod_{j=1}^3 s_{ij}}.
\end{equation}
where $\lambda_{\text{ssim}}$ and  $\gamma$ are hyperparameters, \(s_{ij}\) denotes the scaling factor of the \(i\)-th Gaussian primitive along the \(j\)-th axis (\(j \in \{x,y,z\}\)), and \(\prod_{j=1}^3 s_{ij}\) is the product of the scaling factors for Gaussian \(i\), proportional to its volume.
After obtaining optimized 3DGS parameters, we introduce a joint loss function that combines 3D geometric consistency $L_{\text{GS-point}}$ between positions of optimized Gaussian primitives $P_{\text{GS}}$ and reconstructed points $P_{\text{rec}}$ , and 2D image fidelity $L_{\text{GS-image}}$ between rendered images and ground-truth images :
\begin{equation}
L_{\text{GS-point}} = {\frac{1}{|P_{\text{GS}}|}\sum_{p\in P_{\text{GS}}} \min_{q\in P_{\text{rec}}} \|p-q\|_2^2}+ {\frac{1}{|P_{\text{rec}}|}\sum_{q\in P_{\text{rec}}} \min_{p\in P_{\text{GS}}} \|q-p\|_2^2}
\end{equation}
\begin{equation}
    L_{\text{GS-image}} = (1 - \lambda)L_1 + \lambda L_{\text{D-SSIM}}
\end{equation} 
The joint optimization objective of the 3DGS branch becomes:
\begin{equation}
L_{\text{GS-branch}} =\alpha \cdot L_{\text{GS-image}} + \beta \cdot L_{\text{GS-point}} 
\end{equation} 
Finally, The total loss of second stage is:
\begin{equation}
L_{\text{stage2}} =L_{\text{stage1}}  + L_{\text{GS-branch}}  
\end{equation} 

\section{Experimental Setups}
\label{sec:setups}
We pre-train our model on a subset of multi-view images from SUN RGB-D~\citep{song2015sun} and fine-tune our backbone on downstream tasks on indoor 3D object detection datasets, SUN RGB-D~\citep{song2015sun} and ScanNetV2~\citep{dai2017scannet}.
Following~\citep{pang2022masked}~\citep{gwak2020generative}, the encoder-decoder branch is built upon standard ViT backbones. 
For the inputs to the two branches, the point cloud is down-sampled to 2,048 points, while each 256 × 352 images are divided into uniform patches of size 16 × 16. A masking ratio of 60\% is applied to the point cloud patches.
 \par
During the first pre-training stage, the model is trained for 400 epochs using the AdamW optimizer~\citep{loshchilov2017decoupled} with an initial learning rate of $1e^{-3}$ and a weight decay of 0.05. In the second stage, consistent with the 3DGS branch, we observe that training for one epoch is sufficient. To reconstruct 3D scenes from input images and reconstructed points, we use Scaffold-GS~\citep{lu2024scaffold}, which combining sparse anchors and dynamic refinement strategies and achieving state-of-the-art rendering quality and less computational and storage costs.
For finetuning the downstream task, we mainly follow the settings of~\citep{chen2023pimae}.

\section{Results}
\label{sec:results}

\subsection{Pre-train Results}
\begin{figure}[ht]
    \centering
   
    \begin{minipage}[b]{0.49\linewidth}
        \includegraphics[width=\linewidth]{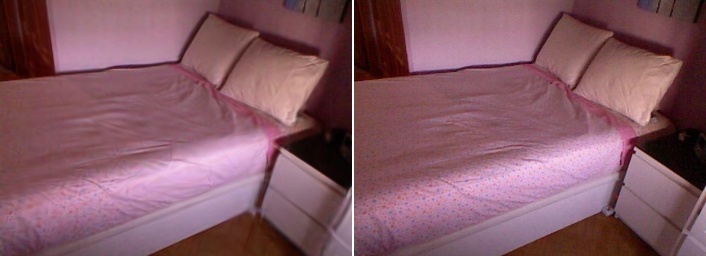}
        \label{fig:render1}
    \end{minipage}
    \hfill
    \begin{minipage}[b]{0.49\linewidth}
        \includegraphics[width=\linewidth]{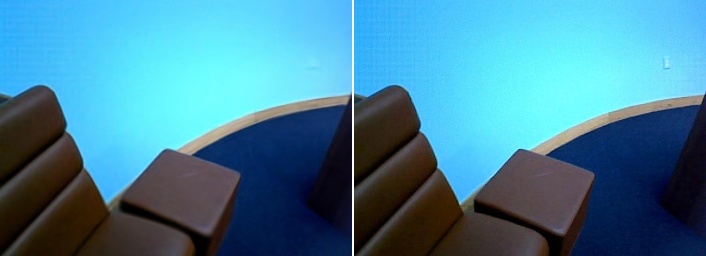}
        \label{fig:render2}
    \end{minipage}

    \begin{minipage}[b]{0.49\linewidth}
        \includegraphics[width=\linewidth]{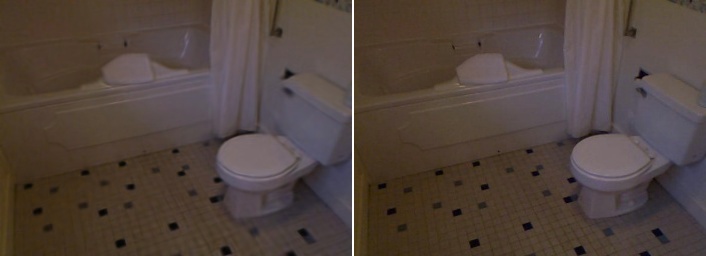}
        \label{fig:render3}
    \end{minipage}
    \hfill
    \begin{minipage}[b]{0.49\linewidth}
        \includegraphics[width=\linewidth]{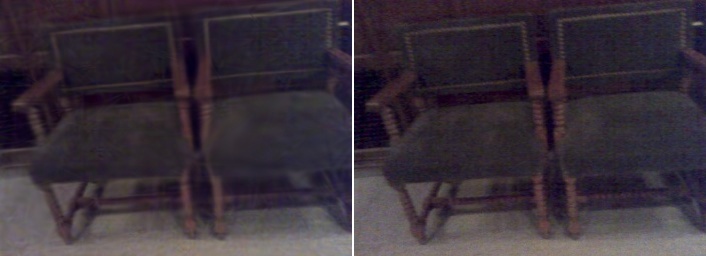}
        \label{fig:render4}
    \end{minipage}

    \begin{minipage}[b]{0.49\linewidth}
        \includegraphics[width=\linewidth]{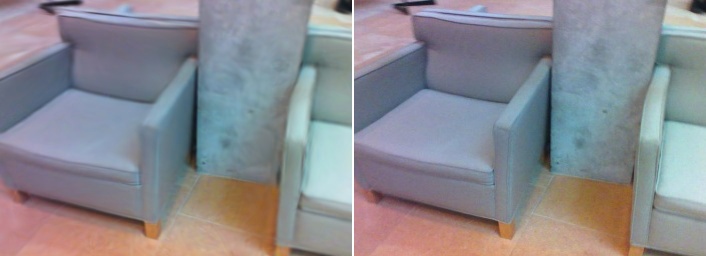}
        \label{fig:render5}
    \end{minipage}
    \hfill
    \begin{minipage}[b]{0.49\linewidth}
        \includegraphics[width=\linewidth]{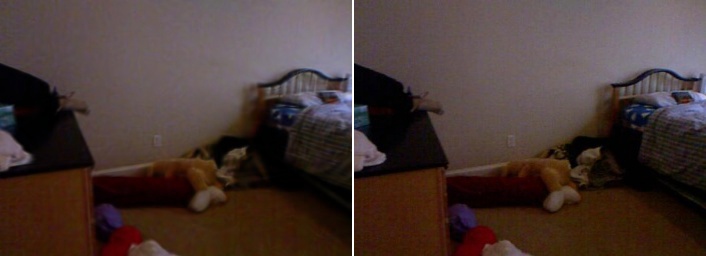}
        \label{fig:render6}
    \end{minipage}
    \caption{3DGS rendering results while pre-training. For each sub-figure, the left one presents the rendering outputs, while the right shows the ground-truth.}
     \label{fig:rendering results}
\end{figure}

\begin{figure}[ht]
    \centering
    
    \begin{minipage}[b]{0.28\linewidth}
            \centering
            \includegraphics[width=\linewidth]{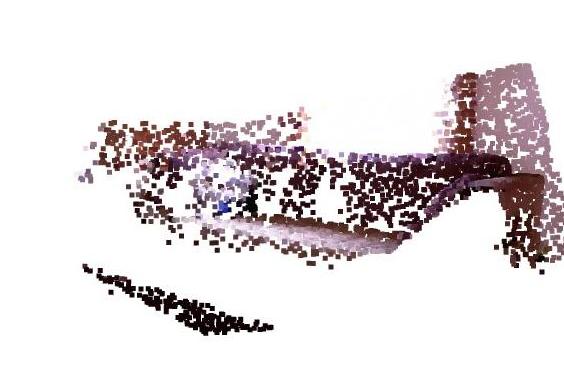}
            \label{fig:points1}
    \end{minipage}
    \hfill
    \begin{minipage}[b]{0.28\linewidth}
            \centering
            \includegraphics[width=\linewidth]{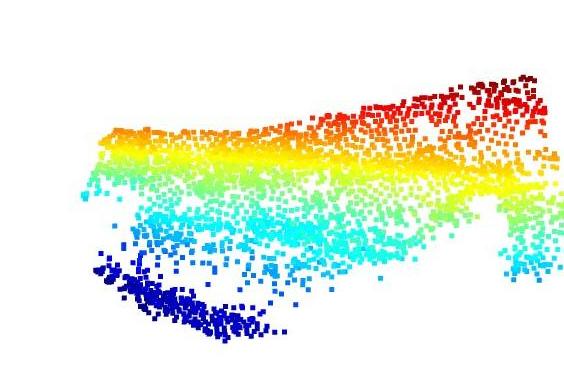}
            \label{fig:points2}
    \end{minipage}
    \hfill
    \begin{minipage}[b]{0.28\linewidth}
            \centering
            \includegraphics[width=\linewidth]{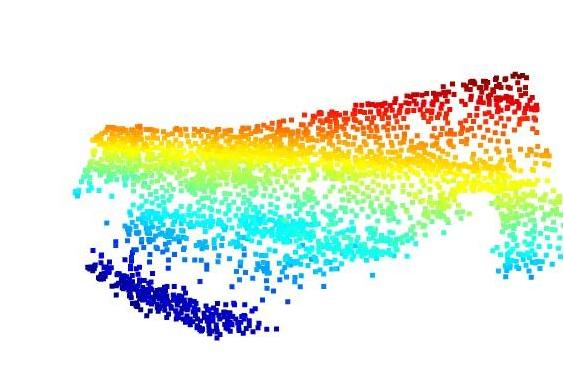}
            \label{fig:points3}
    \end{minipage}

    \begin{minipage}[b]{0.3\linewidth}
            \begin{subfigure}[b]{\linewidth}
            \centering
            \includegraphics[width=\linewidth]{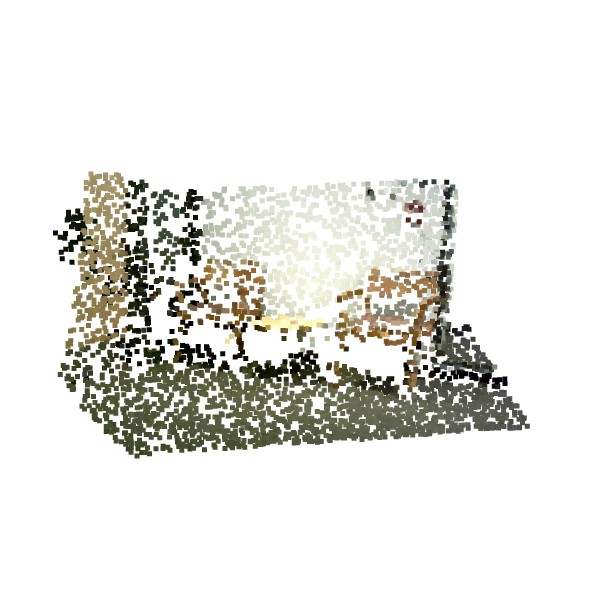}
            \caption{Ground Truth}
            \label{fig:points4}
            \end{subfigure}
    \end{minipage}
    \hfill
    \begin{minipage}[b]{0.3\linewidth}
     \begin{subfigure}[b]{\linewidth}
            \centering
            \includegraphics[width=\linewidth]{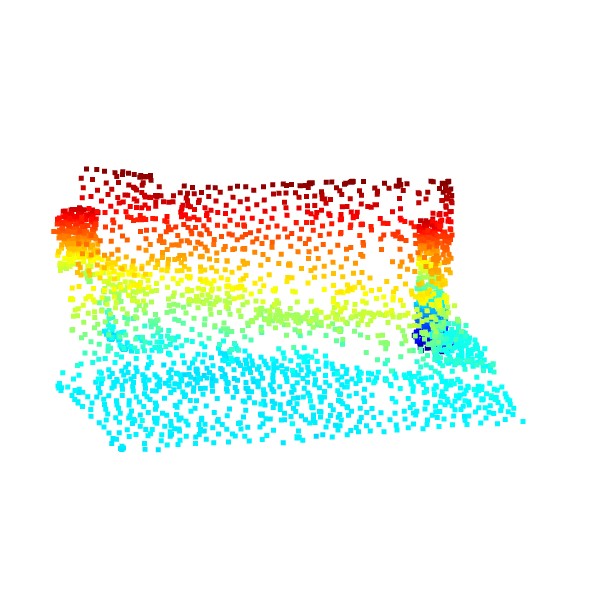}
            \caption{Reconstruction}
            \label{fig:points5}
            \end{subfigure}
    \end{minipage}
    \hfill
    \begin{minipage}[b]{0.3\linewidth}
     \begin{subfigure}[b]{\linewidth}
            \centering
            \includegraphics[width=\linewidth]{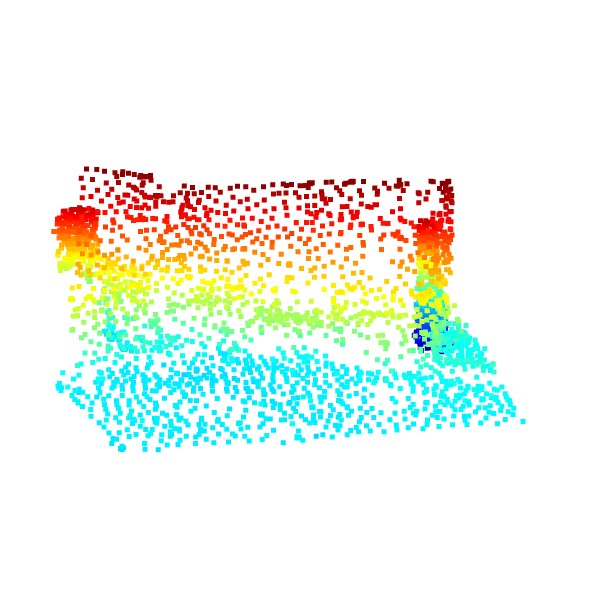}
            \caption{3DGS primitives}
            \label{fig:pointsr6}
        \end{subfigure}
    \end{minipage}
    \caption{Visualization of the reconstructed output and the corresponding Gaussian primitives of the point branch.}
    \label{fig:reconstruction results}
\end{figure}

After pre-training on the subset of multi-view images from SUN RGB-D~\citep{song2015sun}, the model acquires robust feature representation capabilities that capture the geometric and semantic information of scenes. The visualization of the reconstructed output and the corresponding Gaussian points of the point branch is presented in Figure~\ref{fig:reconstruction results}, demonstrating precise alignment between the positions of 3DGS primitives and the original input point cloud. For the image branch, the rendering results of 3DGS are shown in Figure~\ref{fig:rendering results} with an average PSNR of 26.4, surpassing~\citep{zhu2023ponderv2}, which is based on volume rendering.
These results confirm that the 2D renderings and 3D point distributions of 3DGS provide geometrically accurate and semantically rich supervision signals for transferable representations learning.
\begin{table*}[ht]
\centering
\caption{3D object detection per-class average precision on SUN RGB-D with 3D IoU thresholds of 0.25 ($AP_{25}$).}
\label{table:ap25 category}
\resizebox{\linewidth}{!}{
\begin{tabular}{l c c c c c c c c c c c}
\hline 
\textbf{Method} & \textbf{mAP@0.25} & \textbf{bathtub} & \textbf{bed} & \textbf{bookshelf} & \textbf{chair} & \textbf{desk} & \textbf{dresser} & \textbf{nightstand} & \textbf{sofa} & \textbf{table} & \textbf{toilet } \\
\hline
\textbf{3DETR~\cite{misra2021end}}  & 58.0  & 76.3  & 84.2  & 29.3  & 67.4  & 28.8  & 32.4 & 62.6  & 59.8  & 49.2  & 90.1  \\
\midrule
\multirow{1}{*}{\textbf{Ours+3DETR}} &  59.2 & 78.4 & 85.2 & 32.4 & 69.5 & 29.7 & 30.3 & 62.1 & 60.0 & 51.6 &  92.9\\

\textit{Improments} & \textcolor{red}{+1.2\%} & \textcolor{red}{+2.1\%} & \textcolor{red}{+1.0\%} & \textcolor{red}{+3.1\%} & \textcolor{red}{+2.1\%} & \textcolor{red}{+0.9\%} & \textcolor{blue}{-2.1\%} & \textcolor{blue}{-0.5\%} & \textcolor{red}{+0.2\%} & \textcolor{red}{+2.4\%} & \textcolor{red}{+2.8\%} \\
\hline
\end{tabular}
}
\end{table*}

\begin{table*}[ht]
\centering
\caption{3D object detection per-class average precision on SUN RGB-D with 3D IoU thresholds of 0.5 ($AP_{50}$).}
\label{table:ap50 category}
\resizebox{\linewidth}{!}{
\begin{tabular}{l c c c c c c c c c c c}
\hline 
\textbf{Method} & \textbf{mAP@0.50} & \textbf{bathtub} & \textbf{bed} & \textbf{bookshelf} & \textbf{chair} & \textbf{desk} & \textbf{dresser} & \textbf{nightstand} & \textbf{sofa} & \textbf{table} & \textbf{toilet } \\
\hline
\textbf{3DETR~\cite{misra2021end}}  & 30.3  & 39.0   & 51.6   & 3.8   & 38.3   & 5.9   & 17.3   & 27.6   & 37.8   & 18.4   & 63.5    \\ \midrule

\multirow{1}{*}{\textbf{Ours+3DETR}} &  33.5 & 49.2 & 55.1 & 3.8 & 42.1 & 6.9 & 16.4 & 33.5 & 37.8 & 19.7 & 70.2\\

\textit{Improvements} & \textcolor{red}{+3.2\%} & \textcolor{red}{+10.2\%} & \textcolor{red}{+4.5\%} & \textcolor{red}{+0.0\%} & \textcolor{red}{+3.8\%} & \textcolor{red}{+1.0\%} & \textcolor{blue}{-0.9\%} & \textcolor{red}{+5.9\%} & \textcolor{red}{+0.0\%} & \textcolor{red}{+1.3\%} & \textcolor{red}{+6.7\%} \\
\hline
\end{tabular}
}
\end{table*}

\subsection{Downstream Results}
\begin{table}[ht]
\caption{3D object detection results on SUN RGB-D and ScanNetV2. We adopt the average precision (AP) with IoU thresholds of 0.25 and 0.5 for the evaluation metrics.}
\centering
\scalebox{1}{
\begin{tabular}{l|cccc}
\toprule
\multirow{2}{*}{Methods}  & \multicolumn{2}{c}{SUN RGB-D} & \multicolumn{2}{c}{ScanNetV2} \\
&$AP_{25}$ & $AP_{50}$ & $AP_{25}$ & $AP_{50}$ \\
\midrule
 \rowcolor{C7!50}    DSS~\cite{song2016deep}     & 42.1   &  -    & 15.2  & 6.8   \\
 PointFusion~\cite{xu2018pointfusion}   & 45.4   &  -   &   -   &   -  \\
  \rowcolor{C7!50}  3D-SIS~\cite{hou20193d}        &  -  &   -   & 40.2    &  22.5    \\
   VoteNet~\cite{qi2019deep}        &  57.7  &  32.9    &   58.6   & 33.5    \\
   \rowcolor{C7!50} 3DETR~\cite{misra2021end}     &  58.0  &  30.3    &   62.1  &  37.9  \\
   PiMAE + 3DETR~\cite{chen2023pimae}     &  \textbf{59.4}  &   \underline{33.2}    &   \underline{ 62.6}  &   \underline{39.4}  \\
\midrule
\rowcolor{C7!50} Ours+3DETR  &  \underline{59.2} & \textbf{ 33.5}  &   \textbf{62.9}  &  \textbf{43.3}     \\ \bottomrule
\end{tabular}%
}
\label{table:3d object detection}
\end{table}

We transfer the pre-trained models to the 3DETR~\citep{misra2021end} architecture, leveraging geometric priors learned from large-scale indoor-scene datasets to improve detection accuracy on 3D object detection tasks. 
To evaluate the effectiveness of our method, we employ two benchmark datasets: SUN RGB-D~\citep{song2015sun} and ScanNetV2~\citep{dai2017scannet}. 
These datasets offer comprehensive multi-modal indoor scene data, including synchronized RGB-D images,  point clouds, and 3D bounding box annotations. 
In alignment with~\citep{qi2019deep}, we adopt average precision (AP) with IoU thresholds set to $0.25$ and $0.5$ as our metrics for 3D object detection performance.
We report our performance based on 3DETR, which employs a Transformer architecture for end-to-end 3D detection, eliminating dependency on hand-crafted components. 
As shown in Table~\ref{table:3d object detection}, our model substantially outperforms the 3DETR baseline, boosting $AP_{50}$ of $1.2\%$ and $5.4\%$ for SUN RGB-D~\citep{song2015sun} and ScanNetV2~\citep{dai2017scannet}, respectively. 
Our method outperforms the model pre-trained with PiMAE~\citep{chen2023pimae} by $+0.3\%~ AP_{50}$ on SUN RGB-D and $+3.9\%~AP_{50}$  on ScanNetV2, and maintains competitive $AP_{25}$ of  $59.2\%$ and $62.9\%$ on both datasets. 
This demonstrates enhanced capability in predicting geometrically accurate bounding boxes after adding the multi-modal supervision of 3DGS.
Visualization results are shown in Figure~\ref{fig:detection vis}.
Table~\ref{table:ap25 category} and~\ref{table:ap50 category} show the $AP_{25}$ and $AP_{50}$  scores for each category in the SUN RGB-D dataset. 
Our method notably enhances the detection accuracy of the baseline  3DETR model, improving or maintaining detection performance in 8 out of the 10 categories in  $AP_{25}$ and 8 out of the 10 categories in  $AP_{50}$.

\begin{figure}[ht]
    \centering
    \begin{minipage}[b]{0.49\linewidth}
        \includegraphics[width=\linewidth]{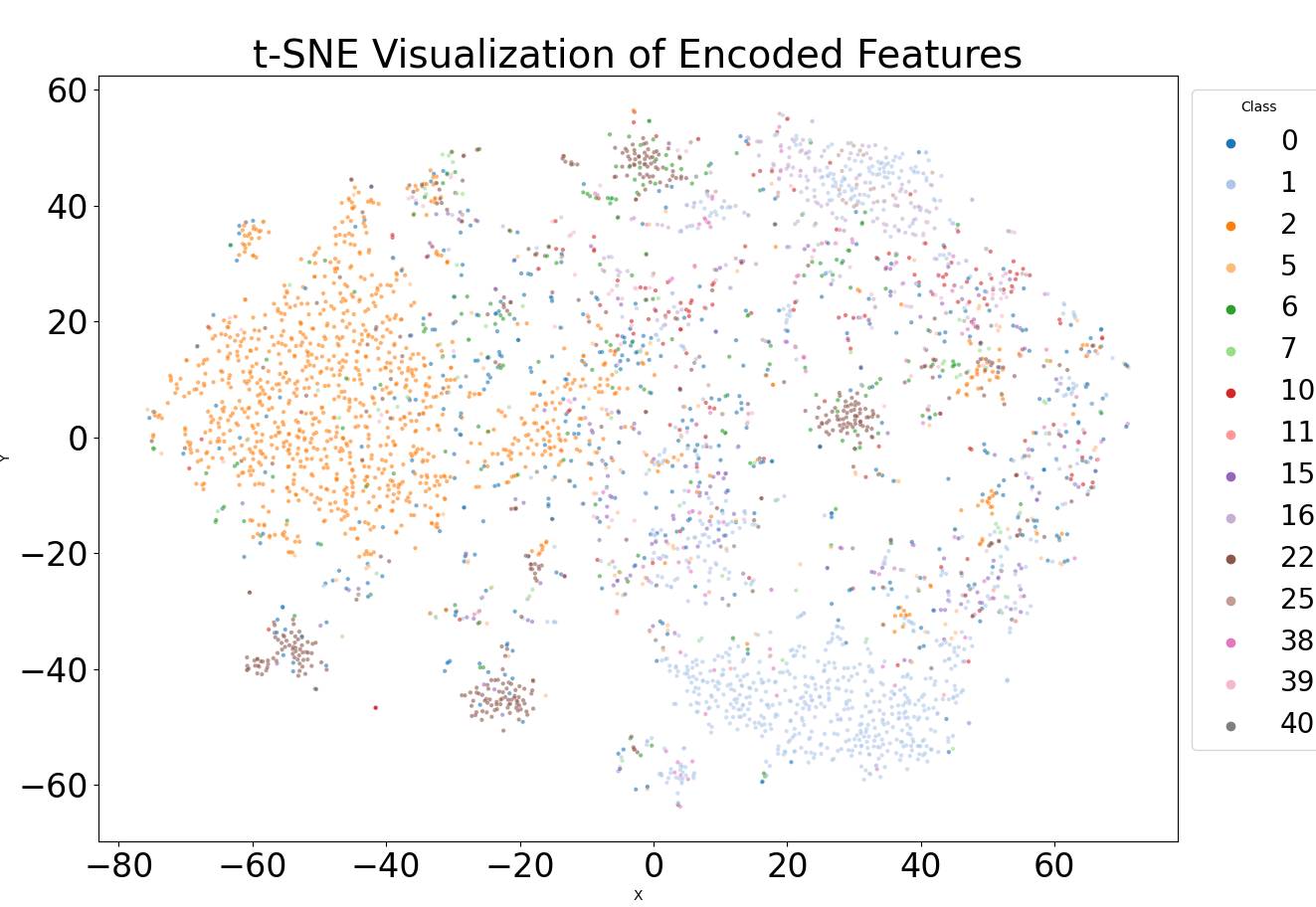}
        \label{fig:tsne1}
    \end{minipage}
    \hfill
    \begin{minipage}[b]{0.49\linewidth}
        \includegraphics[width=\linewidth]{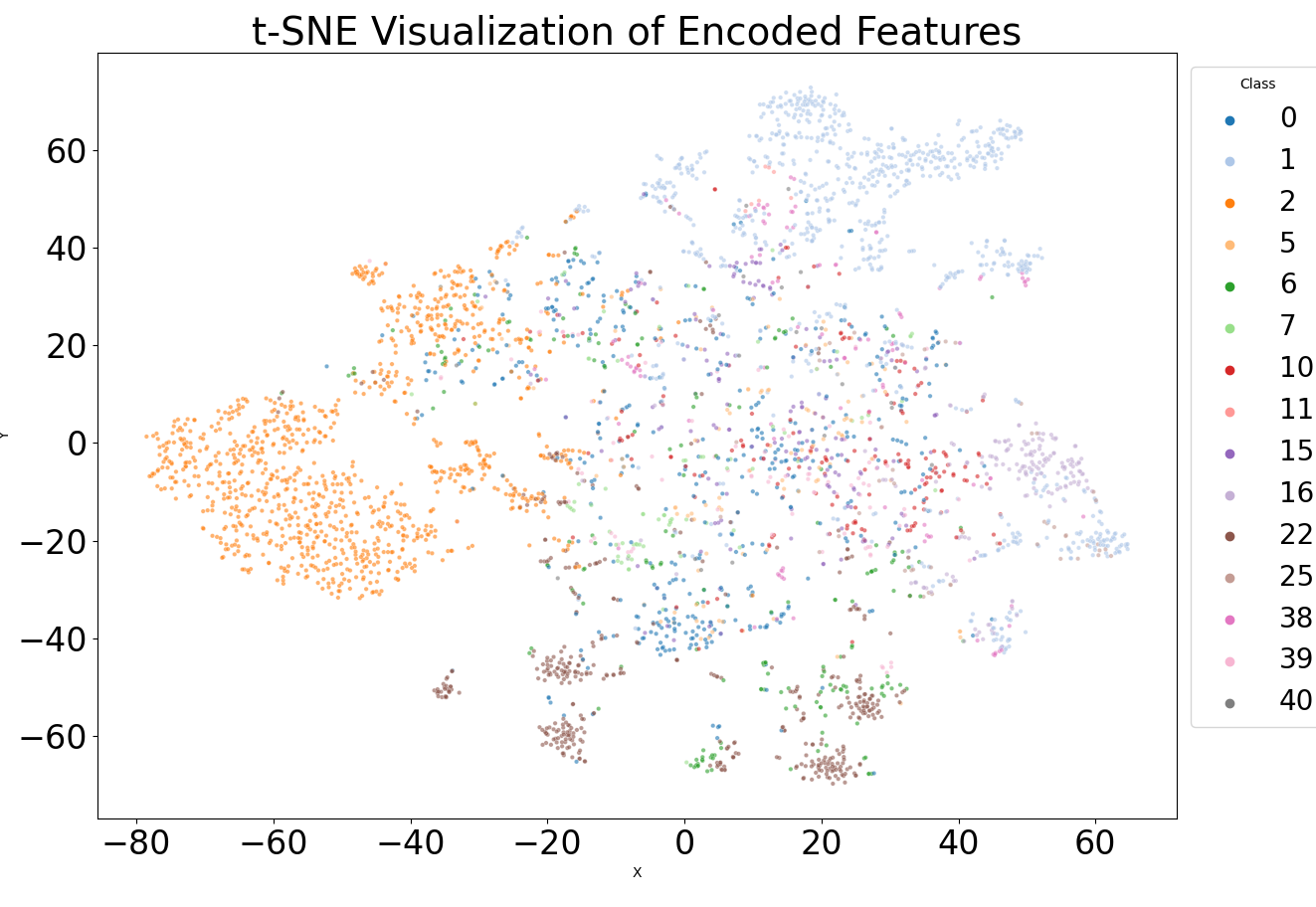}
        \label{fig:tsne2}
    \end{minipage}
    \caption{t-SNE~\citep{van2008visualizing} results of extracted point cloud features on ScanNetV2~\citep{dai2017scannet}. Different colors represent different object categories. The left one shows the feature distribution before pre-training, where class clusters are loosely separated.  The right one demonstrates the enhanced clustering and discriminative capability after training.  }
    \label{fig:tsne results}
\end{figure}

Furthermore, we visualize the t-SNE~\citep{van2008visualizing} embeddings of encoded features on the ScanNetV2 dataset in Figure~\ref{fig:tsne results}. 
Before pre-training, the raw features exhibit poor class separability, with overlapping clusters across categories. In contrast, post-training, the features demonstrate improved clustering, where distinct object classes form compact regions. 

\begin{table}[htbp]
\centering
\begin{minipage}[t]{0.48\linewidth}
\caption{Comparisons of implements of 3DGS modules.}
\centering
\label{table:ablation1}
\tabcolsep=0.3cm
\resizebox{\linewidth}{!}{
\begin{tabular}{ccc|cc|cc}
\toprule
\multicolumn{3}{c|}{Modules} & \multicolumn{2}{c}{SUN RGB-D} & \multicolumn{2}{c}{ScanNetV2} \\
Base & GS\_PC & GS\_IMG & $AP_{25}$ & $AP_{50}$ & $AP_{25}$ & $AP_{50}$ \\ 
\midrule
\rowcolor{C7!50} $\checkmark$ & $\checkmark$ & $\checkmark$ & 59.2 & \textbf{33.5} & 62.9 & \textbf{43.3} \\
\hline
$\checkmark$ & $\checkmark$ & & 59.3 & \underline{32.3} & 63.0 & 42.0 \\
\hline
\rowcolor{C7!50} $\checkmark$ & & $\checkmark$ & 59.4 & 31.0 & \textbf{63.3} & 40.7 \\
\hline
 & $\checkmark$ & $\checkmark$ & 58.7 & 30.8 & 61.9 & 40.2 \\
\hline
\rowcolor{C7!50} & $\checkmark$ & & \underline{59.5} & 31.4 & 62.7 & \underline{42.5} \\
\hline
 & & $\checkmark$ & \textbf{59.6} & 32.0 & \underline{63.0} & 40.6 \\ 
\bottomrule
\end{tabular}}
\end{minipage}
\hfill
\begin{minipage}[t]{0.48\linewidth}
\caption{Effective of different modalities branches.}  
\centering
\label{table:ablation2}
\tabcolsep=0.3cm
\scalebox{1}{
\begin{tabular}{cc|cccc}
\toprule
\multicolumn{2}{c|}{Branches}  & \multicolumn{2}{c}{SUN RGB-D} & \multicolumn{2}{c}{ScanNetV2} \\
PC & IMG & $AP_{25}$ & $AP_{50}$ & $AP_{25}$ & $AP_{50}$ \\
\midrule
 \rowcolor{C7!50} \checkmark & \checkmark & 59.6 & 31.4 & \textbf{62.6} & 42.4 \\
\hline
 \checkmark &  & 59.5 & \textbf{32.4} & 62.5 & \textbf{42.9} \\
\hline
 \rowcolor{C7!50} & \checkmark & \textbf{59.9} & \textbf{32.4} & \textbf{62.6} & 40.4 \\
\bottomrule
\end{tabular}}
\end{minipage}
\end{table}
\begin{table}[htp]
\caption{3D object detection results on SUN RGB-D and ScanNetV2  of different data scales.}
\centering
\tabcolsep=0.3cm
\scalebox{1}{
\begin{tabular}{c|cccc}
\toprule
\multirow{2}{*}{Ratio(\%)}  & \multicolumn{2}{c}{SUN RGB-D} & \multicolumn{2}{c}{ScanNetV2} \\
&$AP_{25}$ & $AP_{50}$ & $AP_{25}$ & $AP_{50}$ \\
\midrule
 \rowcolor{C7!50}    10      &  19.3  &  1.4    & 32.4  & 11.6   \\
 20   & 40.7   &  7.2   &   46.5   &   21.6  \\
  \rowcolor{C7!50}  50        &  56.8  &   29.2   &   59.3  &  38.8   \\
   70        &  59.1    &  31.9       &  61.8   & 39.5    \\
\rowcolor{C7!50}  100        &  59.2  &   33.5   &   62.9  &  43.3   \\
\bottomrule
 
\end{tabular}%
}
\label{table:subset}
\end{table}
\subsection{Ablation Study}
To evaluate the effectiveness of our proposed methods, we conduct ablation studies.  Specifically, we perform two types of ablations.
First, module ablation involves starting from the baseline model that uses combined 3DGS image and point cloud supervision, then selectively removing the additional loss modules applied to the 2D and 3D branches, respectively.
Second, we carry out branch ablation experiments in which only the 2D loss branch or only the 3D loss branch is used independently.
It evaluates complementary effects of the 2D and 3D supervisory signals within the two-branch framework. \par
\textbf{Ablations of 3DGS modules.} As shown in Table~\ref{table:ablation1}, our ablation studies indicate that incorporating only 2D supervision in 3DGS achieves the highest AP\textsubscript{25} of 63.3\%, but yields a lower AP\textsubscript{50}.
Conversely, models pre-trained with only point cloud supervision perform better in terms of AP\textsubscript{50}, but demonstrate inferior AP\textsubscript{25} compared to those trained with only 2D supervision.\par
These results can be attributed to the characteristics of the different supervisory signals.
The 3DGS point cloud supervision directly constrains the 3D bounding box geometry, enabling the model to optimize spatial precision, which translates into higher AP\textsubscript{50}, a metric that demands strict overlap criteria between predictions and ground truth. 
However, point clouds tend to be sparse and incomplete, limiting the model’s ability to robustly detect all object instances, especially under occlusion or in cluttered environments.
This sparsity can reduce recall and thus AP\textsubscript{25}.
Conversely, 2D image supervision through rendering provides rich semantic priors through texture and color, enhancing the model’s ability to recognize and classify targets when spatial information is ambiguous.
This helps the model focus on the presence of objects from a semantic viewpoint, hence boosting AP\textsubscript{25}, which tolerates coarser localization.
Nevertheless, it lacks direct spatial constraints, limiting the fine-grained bounding box accuracy required to raise AP\textsubscript{50}.

\textbf{Ablations of 3DGS branches.}
The branch ablation Table~\ref{table:ablation2} confirms the modality biases that point cloud supervision excels at geometric precision (AP\textsubscript{50}) while image supervision enhances semantic recall (AP\textsubscript{25}). Crucially, simply combining branches without cross-modal reconstruction degrades AP\textsubscript{50} on both datasets and worse than either isolated branch. This evidence shows that unstructured fusion amplifies modality conflicts. The full model (Table~\ref{table:ablation1}) refines this via cross reconstruction, which contains geometric-semantic alignment and structured interaction for effective multimodal fusion.

\textbf{Analysis of data efficiency}
We evaluate the object detection performance of the  model under different training data scales (Table~\ref{table:subset}). The model demonstrates  robustness  when fine-tuned on 70\% of data, which achieves  59.1\% AP\textsubscript{25} on SUN RGB-D~\citep{song2015sun} and 61.8\% AP\textsubscript{25} on ScanNetV2~\citep{dai2017scannet}, 0.1\% and 1.1\% below full-data performance. This robustness stems from the pre-trained encoder's ability to preserve structural priors. However, stricter localization requirements AP\textsubscript{50} (31.9\% AP\textsubscript{50} on SUN RGB-D and 39.5\% AP\textsubscript{50} on ScanNetV2) expose the model's sensitivity to data reduction. The representation learning of fine-grained component-level geometry still requires sufficient supervision signals

\section{Conclusion}
\label{sec:conclusion}
Gaussian2Scene introduces a self-supervised learning method on scene-level point clouds, rethinking how point cloud pre-training interacts with  3D Gaussian Splatting.
We leverage explicit and computationally efficient Gaussian primitives to establish a more direct and accurate connection to 2D rendering and 3D geometry.   
This method employs a progressive two-stage, cross-modal architecture to bridge the gap between 2D and 3D modalities.      
Initially, it learns scene representations through cross-modal masked autoencoding.
Then, it enforces geometric consistency by supervising the learning process with reconstructed Gaussian positions and rendered images.
\bibliographystyle{unsrtnat}
\bibliography{ref}       

\begin{thebibliography}{43}
\providecommand{\natexlab}[1]{#1}
\providecommand{\url}[1]{\texttt{#1}}
\expandafter\ifx\csname urlstyle\endcsname\relax
  \providecommand{\doi}[1]{doi: #1}\else
  \providecommand{\doi}{doi: \begingroup \urlstyle{rm}\Url}\fi

\bibitem[Fei et~al.(2022)Fei, Yang, Chen, Li, Li, Ma, Hu, and Ma]{fei2022comprehensive}
Ben Fei, Weidong Yang, Wen-Ming Chen, Zhijun Li, Yikang Li, Tao Ma, Xing Hu, and Lipeng Ma.
\newblock Comprehensive review of deep learning-based 3d point cloud completion processing and analysis.
\newblock \emph{IEEE Transactions on Intelligent Transportation Systems}, 23\penalty0 (12):\penalty0 22862--22883, 2022.

\bibitem[Ding et~al.(2024)Ding, Zhao, Zhang, Song, Zhang, Huang, Yang, and Wang]{ding2024quar}
Pengxiang Ding, Han Zhao, Wenjie Zhang, Wenxuan Song, Min Zhang, Siteng Huang, Ningxi Yang, and Donglin Wang.
\newblock Quar-vla: Vision-language-action model for quadruped robots.
\newblock In \emph{European Conference on Computer Vision}, pages 352--367. Springer, 2024.

\bibitem[Long et~al.(2024)Long, Li, Cai, and Dong]{long2024discuss}
Yuxing Long, Xiaoqi Li, Wenzhe Cai, and Hao Dong.
\newblock Discuss before moving: Visual language navigation via multi-expert discussions.
\newblock In \emph{2024 IEEE International Conference on Robotics and Automation (ICRA)}, pages 17380--17387. IEEE, 2024.

\bibitem[Fei et~al.(2024{\natexlab{a}})Fei, Luo, Yang, Liu, Zhang, and He]{fei2024curriculumformer}
Ben Fei, Tianyue Luo, Weidong Yang, Liwen Liu, Rui Zhang, and Ying He.
\newblock Curriculumformer: Taming curriculum pre-training for enhanced 3-d point cloud understanding.
\newblock \emph{IEEE Transactions on Neural Networks and Learning Systems}, 2024{\natexlab{a}}.

\bibitem[Fei et~al.(2023)Fei, Yang, Liu, Luo, Zhang, Li, and He]{fei2023self}
Ben Fei, Weidong Yang, Liwen Liu, Tianyue Luo, Rui Zhang, Yixuan Li, and Ying He.
\newblock Self-supervised learning for pre-training 3d point clouds: A survey.
\newblock \emph{arXiv preprint arXiv:2305.04691}, 2023.

\bibitem[Fei et~al.(2024{\natexlab{b}})Fei, Liu, Yang, Li, Chen, and Ma]{fei2024parameter}
Ben Fei, Liwen Liu, Weidong Yang, Zhijun Li, Wen-Ming Chen, and Lipeng Ma.
\newblock Parameter efficient point cloud prompt tuning for unified point cloud understanding.
\newblock \emph{IEEE Transactions on Intelligent Vehicles}, 2024{\natexlab{b}}.

\bibitem[Wang et~al.(2021)Wang, Liu, Yue, Lasenby, and Kusner]{wang2021unsupervised}
Hanchen Wang, Qi~Liu, Xiangyu Yue, Joan Lasenby, and Matt~J Kusner.
\newblock Unsupervised point cloud pre-training via occlusion completion.
\newblock In \emph{Proceedings of the IEEE/CVF international conference on computer vision}, pages 9782--9792, 2021.

\bibitem[Yu et~al.(2022)Yu, Tang, Rao, Huang, Zhou, and Lu]{yu2022point}
Xumin Yu, Lulu Tang, Yongming Rao, Tiejun Huang, Jie Zhou, and Jiwen Lu.
\newblock Point-bert: Pre-training 3d point cloud transformers with masked point modeling.
\newblock In \emph{Proceedings of the IEEE/CVF conference on computer vision and pattern recognition}, pages 19313--19322, 2022.

\bibitem[Pang et~al.(2022)Pang, Wang, Tay, Liu, Tian, and Yuan]{pang2022masked}
Yatian Pang, Wenxiao Wang, Francis~EH Tay, Wei Liu, Yonghong Tian, and Li~Yuan.
\newblock Masked autoencoders for point cloud self-supervised learning.
\newblock In \emph{European conference on computer vision}, pages 604--621. Springer, 2022.

\bibitem[Zhang et~al.(2022)Zhang, Guo, Gao, Fang, Zhao, Wang, Qiao, and Li]{zhang2022point}
Renrui Zhang, Ziyu Guo, Peng Gao, Rongyao Fang, Bin Zhao, Dong Wang, Yu~Qiao, and Hongsheng Li.
\newblock Point-m2ae: multi-scale masked autoencoders for hierarchical point cloud pre-training.
\newblock \emph{Advances in neural information processing systems}, 35:\penalty0 27061--27074, 2022.

\bibitem[Xie et~al.(2020)Xie, Gu, Guo, Qi, Guibas, and Litany]{xie2020pointcontrast}
Saining Xie, Jiatao Gu, Demi Guo, Charles~R Qi, Leonidas Guibas, and Or~Litany.
\newblock Pointcontrast: Unsupervised pre-training for 3d point cloud understanding.
\newblock In \emph{Computer Vision--ECCV 2020: 16th European Conference, Glasgow, UK, August 23--28, 2020, Proceedings, Part III 16}, pages 574--591. Springer, 2020.

\bibitem[Afham et~al.(2022)Afham, Dissanayake, Dissanayake, Dharmasiri, Thilakarathna, and Rodrigo]{afham2022crosspoint}
Mohamed Afham, Isuru Dissanayake, Dinithi Dissanayake, Amaya Dharmasiri, Kanchana Thilakarathna, and Ranga Rodrigo.
\newblock Crosspoint: Self-supervised cross-modal contrastive learning for 3d point cloud understanding.
\newblock In \emph{Proceedings of the IEEE/CVF Conference on Computer Vision and Pattern Recognition}, pages 9902--9912, 2022.

\bibitem[Huang et~al.(2021)Huang, Xie, Zhu, and Zhu]{huang2021spatio}
Siyuan Huang, Yichen Xie, Song-Chun Zhu, and Yixin Zhu.
\newblock Spatio-temporal self-supervised representation learning for 3d point clouds.
\newblock \emph{arXiv preprint arXiv:2109.00179}, 2021.

\bibitem[Huang et~al.(2023)Huang, Peng, He, Yang, Zhou, and Ouyang]{huang2023ponder}
Di~Huang, Sida Peng, Tong He, Honghui Yang, Xiaowei Zhou, and Wanli Ouyang.
\newblock Ponder: Point cloud pre-training via neural rendering.
\newblock In \emph{Proceedings of the IEEE/CVF International Conference on Computer Vision}, pages 16089--16098, 2023.

\bibitem[Song et~al.(2024)Song, Zeng, Ren, and Zhang]{song2024city}
Kaiwen Song, Xiaoyi Zeng, Chenqu Ren, and Juyong Zhang.
\newblock City-on-web: real-time neural rendering of large-scale scenes on the web.
\newblock In \emph{European Conference on Computer Vision}, pages 385--402. Springer, 2024.

\bibitem[Kerbl et~al.(2023)Kerbl, Kopanas, Leimk{\"u}hler, and Drettakis]{kerbl20233d}
Bernhard Kerbl, Georgios Kopanas, Thomas Leimk{\"u}hler, and George Drettakis.
\newblock 3d gaussian splatting for real-time radiance field rendering.
\newblock \emph{ACM Trans. Graph.}, 42\penalty0 (4):\penalty0 139--1, 2023.

\bibitem[Fei et~al.(2024{\natexlab{c}})Fei, Xu, Zhang, Zhou, Yang, and He]{fei20243d}
Ben Fei, Jingyi Xu, Rui Zhang, Qingyuan Zhou, Weidong Yang, and Ying He.
\newblock 3d gaussian splatting as new era: A survey.
\newblock \emph{IEEE Transactions on Visualization and Computer Graphics}, 2024{\natexlab{c}}.

\bibitem[Liu et~al.(2025)Liu, Luo, Yang, Xu, Li, Chen, and Fei]{liu2025gs}
Keyi Liu, Yeqi Luo, Weidong Yang, Jingyi Xu, Zhijun Li, Wen-Ming Chen, and Ben Fei.
\newblock Gs-pt: Exploiting 3d gaussian splatting for comprehensive point cloud understanding via self-supervised learning.
\newblock In \emph{ICASSP 2025-2025 IEEE International Conference on Acoustics, Speech and Signal Processing (ICASSP)}, pages 1--5. IEEE, 2025.

\bibitem[Misra et~al.(2021)Misra, Girdhar, and Joulin]{misra2021end}
Ishan Misra, Rohit Girdhar, and Armand Joulin.
\newblock An end-to-end transformer model for 3d object detection.
\newblock In \emph{Proceedings of the IEEE/CVF international conference on computer vision}, pages 2906--2917, 2021.

\bibitem[Song et~al.(2015)Song, Lichtenberg, and Xiao]{song2015sun}
Shuran Song, Samuel~P Lichtenberg, and Jianxiong Xiao.
\newblock Sun rgb-d: A rgb-d scene understanding benchmark suite.
\newblock In \emph{Proceedings of the IEEE conference on computer vision and pattern recognition}, pages 567--576, 2015.

\bibitem[Dai et~al.(2017)Dai, Chang, Savva, Halber, Funkhouser, and Nie{\ss}ner]{dai2017scannet}
Angela Dai, Angel~X Chang, Manolis Savva, Maciej Halber, Thomas Funkhouser, and Matthias Nie{\ss}ner.
\newblock Scannet: Richly-annotated 3d reconstructions of indoor scenes.
\newblock In \emph{Proceedings of the IEEE conference on computer vision and pattern recognition}, pages 5828--5839, 2017.

\bibitem[Mildenhall et~al.(2021)Mildenhall, Srinivasan, Tancik, Barron, Ramamoorthi, and Ng]{mildenhall2021nerf}
Ben Mildenhall, Pratul~P Srinivasan, Matthew Tancik, Jonathan~T Barron, Ravi Ramamoorthi, and Ren Ng.
\newblock Nerf: Representing scenes as neural radiance fields for view synthesis.
\newblock \emph{Communications of the ACM}, 65\penalty0 (1):\penalty0 99--106, 2021.

\bibitem[Gu{\'e}don and Lepetit(2024)]{guedon2024sugar}
Antoine Gu{\'e}don and Vincent Lepetit.
\newblock Sugar: Surface-aligned gaussian splatting for efficient 3d mesh reconstruction and high-quality mesh rendering.
\newblock In \emph{Proceedings of the IEEE/CVF Conference on Computer Vision and Pattern Recognition}, pages 5354--5363, 2024.

\bibitem[Yang et~al.(2024{\natexlab{a}})Yang, Gao, Zhou, Jiao, Zhang, and Jin]{yang2024deformable}
Ziyi Yang, Xinyu Gao, Wen Zhou, Shaohui Jiao, Yuqing Zhang, and Xiaogang Jin.
\newblock Deformable 3d gaussians for high-fidelity monocular dynamic scene reconstruction.
\newblock In \emph{Proceedings of the IEEE/CVF Conference on Computer Vision and Pattern Recognition}, pages 20331--20341, 2024{\natexlab{a}}.

\bibitem[Lin et~al.(2024)Lin, Li, Tang, Liu, Liu, Liu, Lu, Wu, Xu, Yan, et~al.]{lin2024vastgaussian}
Jiaqi Lin, Zhihao Li, Xiao Tang, Jianzhuang Liu, Shiyong Liu, Jiayue Liu, Yangdi Lu, Xiaofei Wu, Songcen Xu, Youliang Yan, et~al.
\newblock Vastgaussian: Vast 3d gaussians for large scene reconstruction.
\newblock In \emph{Proceedings of the IEEE/CVF Conference on Computer Vision and Pattern Recognition}, pages 5166--5175, 2024.

\bibitem[Chen et~al.(2024)Chen, Chen, Zhang, Wang, Yang, Wang, Cai, Yang, Liu, and Lin]{chen2024gaussianeditor}
Yiwen Chen, Zilong Chen, Chi Zhang, Feng Wang, Xiaofeng Yang, Yikai Wang, Zhongang Cai, Lei Yang, Huaping Liu, and Guosheng Lin.
\newblock Gaussianeditor: Swift and controllable 3d editing with gaussian splatting.
\newblock In \emph{Proceedings of the IEEE/CVF Conference on Computer Vision and Pattern Recognition}, pages 21476--21485, 2024.

\bibitem[Liang et~al.(2024)Liang, Yang, Lin, Li, Xu, and Chen]{liang2024luciddreamer}
Yixun Liang, Xin Yang, Jiantao Lin, Haodong Li, Xiaogang Xu, and Yingcong Chen.
\newblock Luciddreamer: Towards high-fidelity text-to-3d generation via interval score matching.
\newblock In \emph{Proceedings of the IEEE/CVF Conference on Computer Vision and Pattern Recognition}, pages 6517--6526, 2024.

\bibitem[Zhou et~al.(2024)Zhou, Lin, Shan, Wang, Sun, and Yang]{zhou2024drivinggaussian}
Xiaoyu Zhou, Zhiwei Lin, Xiaojun Shan, Yongtao Wang, Deqing Sun, and Ming-Hsuan Yang.
\newblock Drivinggaussian: Composite gaussian splatting for surrounding dynamic autonomous driving scenes.
\newblock In \emph{Proceedings of the IEEE/CVF Conference on Computer Vision and Pattern Recognition}, pages 21634--21643, 2024.

\bibitem[Jiang et~al.(2024)Jiang, Shen, Wang, Su, Hong, Zhang, Yu, and Xu]{jiang2024hifi4g}
Yuheng Jiang, Zhehao Shen, Penghao Wang, Zhuo Su, Yu~Hong, Yingliang Zhang, Jingyi Yu, and Lan Xu.
\newblock Hifi4g: High-fidelity human performance rendering via compact gaussian splatting.
\newblock In \emph{Proceedings of the IEEE/CVF Conference on Computer Vision and Pattern Recognition}, pages 19734--19745, 2024.

\bibitem[Zhu et~al.(2023)Zhu, Yang, Wu, Huang, Zhang, He, He, Zhao, Shen, Qiao, and Ouyang]{zhu2023ponderv2}
Haoyi Zhu, Honghui Yang, Xiaoyang Wu, Di~Huang, Sha Zhang, Xianglong He, Tong He, Hengshuang Zhao, Chunhua Shen, Yu~Qiao, and Wanli Ouyang.
\newblock Ponderv2: Pave the way for 3d foundation model with a universal pre-training paradigm.
\newblock \emph{arXiv preprint arXiv:2310.08586}, 2023.

\bibitem[Mei et~al.(2024)Mei, Saltori, Ricci, Sebe, Wu, Zhang, and Poiesi]{mei2024unsupervised}
Guofeng Mei, Cristiano Saltori, Elisa Ricci, Nicu Sebe, Qiang Wu, Jian Zhang, and Fabio Poiesi.
\newblock Unsupervised point cloud representation learning by clustering and neural rendering.
\newblock \emph{International Journal of Computer Vision}, 132\penalty0 (8):\penalty0 3251--3269, 2024.

\bibitem[Yang et~al.(2024{\natexlab{b}})Yang, Zhang, Huang, Wu, Zhu, He, Tang, Zhao, Qiu, Lin, et~al.]{yang2024unipad}
Honghui Yang, Sha Zhang, Di~Huang, Xiaoyang Wu, Haoyi Zhu, Tong He, Shixiang Tang, Hengshuang Zhao, Qibo Qiu, Binbin Lin, et~al.
\newblock Unipad: A universal pre-training paradigm for autonomous driving.
\newblock In \emph{Proceedings of the IEEE/CVF Conference on Computer Vision and Pattern Recognition}, pages 15238--15250, 2024{\natexlab{b}}.

\bibitem[Chen et~al.(2023)Chen, Zhang, Zhang, Wang, Lu, Guo, and Zhang]{chen2023pimae}
Anthony Chen, Kevin Zhang, Renrui Zhang, Zihan Wang, Yuheng Lu, Yandong Guo, and Shanghang Zhang.
\newblock Pimae: Point cloud and image interactive masked autoencoders for 3d object detection.
\newblock In \emph{Proceedings of the IEEE/CVF Conference on Computer Vision and Pattern Recognition}, 2023.

\bibitem[Dosovitskiy et~al.(2020)Dosovitskiy, Beyer, Kolesnikov, Weissenborn, Zhai, Unterthiner, Dehghani, Minderer, Heigold, Gelly, et~al.]{dosovitskiy2020image}
Alexey Dosovitskiy, Lucas Beyer, Alexander Kolesnikov, Dirk Weissenborn, Xiaohua Zhai, Thomas Unterthiner, Mostafa Dehghani, Matthias Minderer, Georg Heigold, Sylvain Gelly, et~al.
\newblock An image is worth 16x16 words: Transformers for image recognition at scale.
\newblock \emph{arXiv preprint arXiv:2010.11929}, 2020.

\bibitem[Wang et~al.(2004)Wang, Bovik, Sheikh, and Simoncelli]{wang2004image}
Zhou Wang, Alan~C Bovik, Hamid~R Sheikh, and Eero~P Simoncelli.
\newblock Image quality assessment: from error visibility to structural similarity.
\newblock \emph{IEEE transactions on image processing}, 13\penalty0 (4):\penalty0 600--612, 2004.

\bibitem[Gwak et~al.(2020)Gwak, Choy, and Savarese]{gwak2020generative}
JunYoung Gwak, Christopher Choy, and Silvio Savarese.
\newblock Generative sparse detection networks for 3d single-shot object detection.
\newblock In \emph{Computer Vision--ECCV 2020: 16th European Conference, Glasgow, UK, August 23--28, 2020, Proceedings, Part IV 16}, pages 297--313. Springer, 2020.

\bibitem[Loshchilov and Hutter(2017)]{loshchilov2017decoupled}
Ilya Loshchilov and Frank Hutter.
\newblock Decoupled weight decay regularization.
\newblock \emph{arXiv preprint arXiv:1711.05101}, 2017.

\bibitem[Lu et~al.(2024)Lu, Yu, Xu, Xiangli, Wang, Lin, and Dai]{lu2024scaffold}
Tao Lu, Mulin Yu, Linning Xu, Yuanbo Xiangli, Limin Wang, Dahua Lin, and Bo~Dai.
\newblock Scaffold-gs: Structured 3d gaussians for view-adaptive rendering.
\newblock In \emph{Proceedings of the IEEE/CVF Conference on Computer Vision and Pattern Recognition}, pages 20654--20664, 2024.

\bibitem[Song and Xiao(2016)]{song2016deep}
Shuran Song and Jianxiong Xiao.
\newblock Deep sliding shapes for amodal 3d object detection in rgb-d images.
\newblock In \emph{Proceedings of the IEEE conference on computer vision and pattern recognition}, pages 808--816, 2016.

\bibitem[Xu et~al.(2018)Xu, Anguelov, and Jain]{xu2018pointfusion}
Danfei Xu, Dragomir Anguelov, and Ashesh Jain.
\newblock Pointfusion: Deep sensor fusion for 3d bounding box estimation.
\newblock In \emph{Proceedings of the IEEE conference on computer vision and pattern recognition}, pages 244--253, 2018.

\bibitem[Hou et~al.(2019)Hou, Dai, and Nie{\ss}ner]{hou20193d}
Ji~Hou, Angela Dai, and Matthias Nie{\ss}ner.
\newblock 3d-sis: 3d semantic instance segmentation of rgb-d scans.
\newblock In \emph{Proceedings of the IEEE/CVF conference on computer vision and pattern recognition}, pages 4421--4430, 2019.

\bibitem[Qi et~al.(2019)Qi, Litany, He, and Guibas]{qi2019deep}
Charles~R Qi, Or~Litany, Kaiming He, and Leonidas~J Guibas.
\newblock Deep hough voting for 3d object detection in point clouds.
\newblock In \emph{proceedings of the IEEE/CVF International Conference on Computer Vision}, pages 9277--9286, 2019.

\bibitem[Van~der Maaten and Hinton(2008)]{van2008visualizing}
Laurens Van~der Maaten and Geoffrey Hinton.
\newblock Visualizing data using t-sne.
\newblock \emph{Journal of machine learning research}, 9\penalty0 (11), 2008.

\end{thebibliography}



\end{document}